%% file: behley2019arxiv.tex
\documentclass[10pt,twocolumn,letterpaper]{article}

\usepackage{iccv}
\usepackage{times}
\usepackage{epsfig}
\usepackage{graphicx}
\usepackage{amsmath}
\usepackage{amssymb}
\usepackage{booktabs}
\usepackage{rotating}
\usepackage{multirow}
\usepackage{caption}
\usepackage{xspace}
\usepackage[table, svgnames, dvipsnames]{xcolor}
\usepackage{pifont}% http://ctan.org/pkg/pifont
\usepackage{longtable}
\usepackage[symbol]{footmisc}
\usepackage{siunitx}
\usepackage{makecell}

\usepackage[pagebackref=true,breaklinks=true,letterpaper=true,colorlinks=true,bookmarks=false]{hyperref}

\iccvfinalcopy % *** Uncomment this line for the final submission

\def\iccvPaperID{2944} % *** Enter the CVPR Paper ID here

\newif\ifspgraph

\spgraphtrue

\newcommand{\datasetname}{SemanticKITTI\xspace}

\makeatletter
\renewcommand{\paragraph}{%
  \@startsection{paragraph}{4}%
  {\z@}{1.0ex \@plus 0.5ex \@minus .2ex}{-1em}%
  {\normalfont\normalsize\bfseries}%
}
\makeatother

\begin{document}

%%%%%%%%% TITLE
\title{\datasetname: A Dataset for Semantic Scene Understanding\\ of LiDAR Sequences}

\author{Jens Behley$^{*}$  \hspace{2cm} Martin Garbade$^{*}$ \hspace{2cm} Andres Milioto \hspace{2cm} Jan Quenzel \\ 
Sven Behnke \hspace{2cm} Cyrill Stachniss  \hspace{2cm} Juergen Gall \\
University of Bonn, Germany}

% more intutive macros for set operations
\newcommand{\intersect}[0]{\cap}
\newcommand{\union}[0]{\cup}
\newcommand{\difference}[0]{-}

%% some number spaces
\newcommand{\RR}{\mathbb{R}} %% real numbers
\newcommand{\NN}{\mathbb{N}} %% natural numbers
\newcommand{\ZZ}{\mathbb{Z}} %% integers

\newcommand{\pd}[2]{\frac{\partial #1}{\partial #2}} %% partial derivative

%% gamma function, and digamma(Psi)
\renewcommand{\digamma}[1]{\Psi\left(#1\right)}
\newcommand{\tgamma}[1]{\Gamma\left(#1\right)}

%% some notation for proababilities
\newcommand{\prob}[1][P]{#1}
\newcommand{\probof}[2][P]{\prob[#1]\left(#2\right)}
\newcommand{\expectation}[2]{\mathbf{E}_{#1}\left[#2\right]}
\newcommand{\entropy}[2]{\mathbf{H}_{#1}\left[#2\right]}
\newcommand{\relentropy}[2]{\mathbf{D}\left(#1||#2\right)}

%% some notation for vectors, matricies and sets
\newcommand{\veca}[1]{\vec{#1}}		%% vector with an arrow above
\renewcommand{\vec}[1]{\mathbf{#1}}	%% vectors are bold
\newcommand{\mat}[1]{\mathbf{#1}}  	%% matricies are also bold
\newcommand{\set}[1]{\mathcal{#1}} 	%% sets are denoted by calligraphic letters

\newcommand{\refsec}[1]{Section~\ref{#1}}
\newcommand{\reffig}[1]{Figure~\ref{#1}}
\newcommand{\reftab}[1]{Table~\ref{#1}}
\newcommand{\refeq}[1]{Equation~\ref{#1}}

\newcommand{\todo}[1]{\noindent\textbf{[TODO: #1\hfill]}\newline\noindent}
\newcommand{\JG}[1]{\noindent\textbf{[JG: #1\hfill]}\newline\noindent}

\newcommand{\cmark}{\ding{51}}%
\newcommand{\xmark}{\ding{55}}%

%\maketitle
\newlength{\teaserwidth}
\setlength{\teaserwidth}{0.225\textwidth}
\newlength{\teaserspace}
\setlength{\teaserspace}{0.3cm}

\twocolumn[{%
\renewcommand\twocolumn[1][]{#1}%
\maketitle
\begin{center}
  \vspace{-0.6cm}
  \url{www.semantic-kitti.org}\\
  \vspace{0.4cm}
  \includegraphics[width=\teaserwidth]{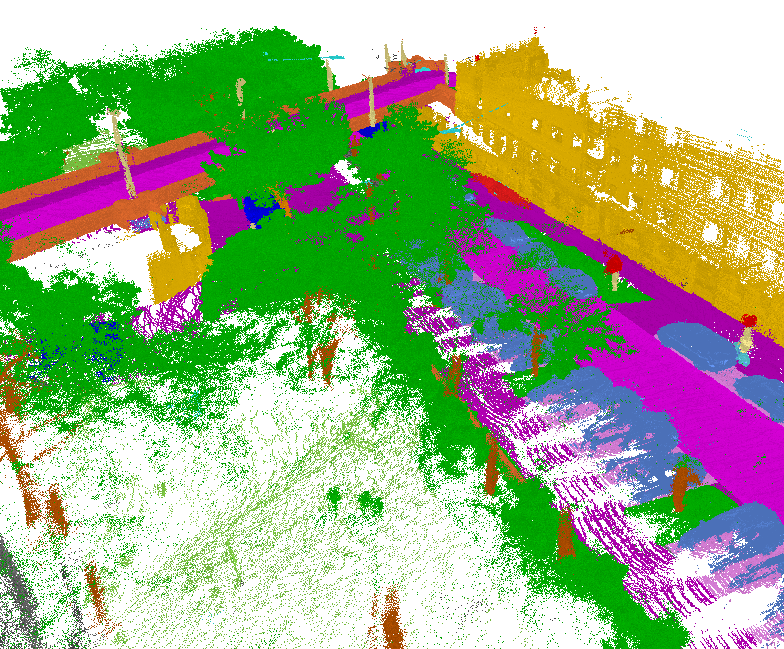}
  \hspace{\teaserspace}
  \includegraphics[width=\teaserwidth]{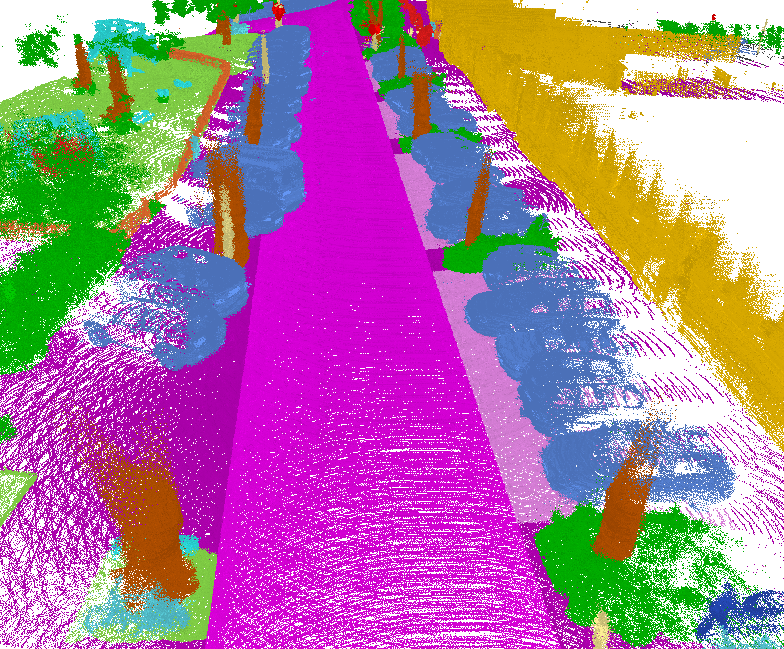}
  \hspace{\teaserspace}
  \includegraphics[width=\teaserwidth]{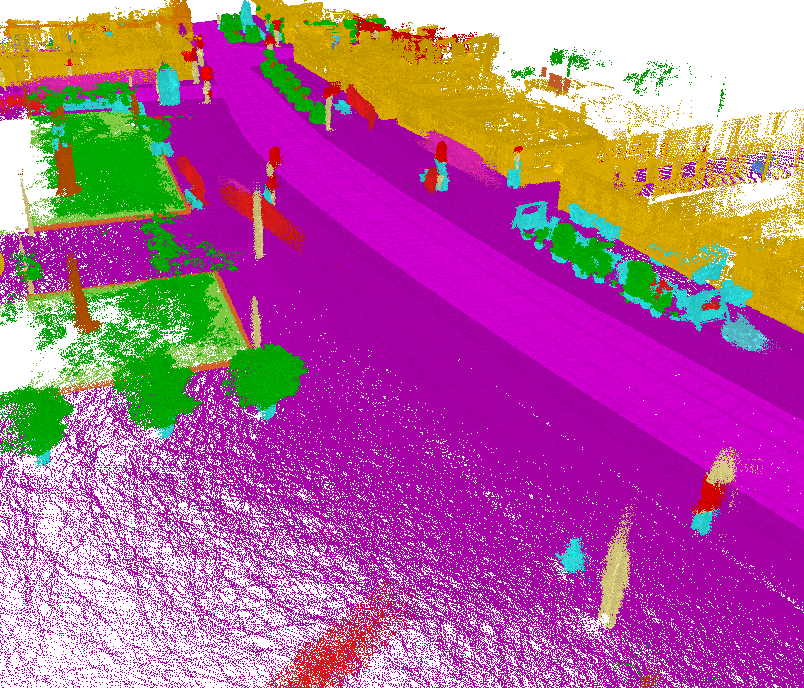}
  \hspace{\teaserspace}
  \includegraphics[width=\teaserwidth]{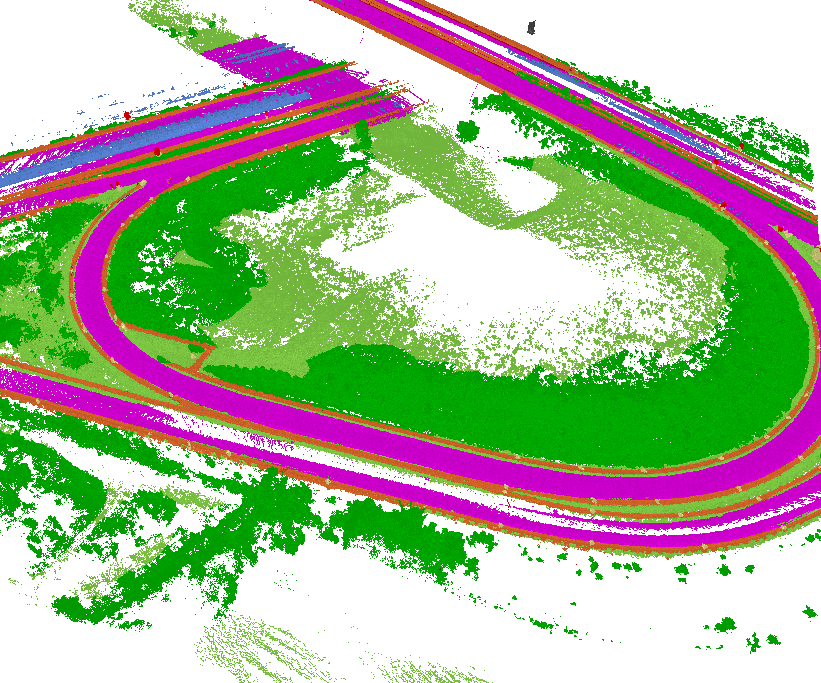}
    \captionof{figure}{Our dataset provides dense annotations for each scan of all sequences from the KITTI Odometry Benchmark \cite{geiger2012cvpr}. 
    Here, we show multiple scans aggregated using pose information estimated by a SLAM approach. 
    }
    \vspace{0.0cm}
\end{center}%
}]

\newcolumntype{L}[1]{>{\raggedright\arraybackslash}p{#1}}
\newcolumntype{C}[1]{>{\centering\arraybackslash}p{#1}}
\newcolumntype{R}[1]{>{\raggedleft\arraybackslash}p{#1}}

\renewcommand{\thefootnote}{\fnsymbol{footnote}}

%%%%%%%%% ABSTRACT
\begin{abstract}
\footnotetext[0]{\vspace{-0.6cm}$^*$ indicates equal contribution} 
Semantic scene understanding is important for various applications.
In particular, self-driving cars need a fine-grained understanding of the surfaces and objects in their vicinity.
Light detection and ranging (LiDAR) provides precise geometric information about the environment and is thus a part of the sensor suites of almost all self-driving cars.
Despite the relevance of semantic scene understanding for this application, there is a lack of a large dataset for this task which is based on an automotive LiDAR.

In this paper, we introduce a large dataset to propel research on laser-based semantic segmentation.  
We annotated all sequences of the KITTI Vision Odometry Benchmark and provide dense point-wise annotations for the complete $360^{o}$ field-of-view of the employed automotive LiDAR.
We propose three benchmark tasks based on this dataset: (i)~semantic segmentation of point clouds using a single scan, (ii)~semantic segmentation using multiple past scans, and (iii)~semantic scene completion, which requires to anticipate the semantic scene in the future. 
We provide baseline experiments and show that there is a need for more sophisticated models to efficiently tackle these tasks. 
Our dataset opens the door for the development of more advanced methods, but also provides plentiful data to investigate new research directions. 
\end{abstract}

\renewcommand*{\thefootnote}{\arabic{footnote}}

%%%%%%%%% BODY TEXT
\section{Introduction}

Semantic scene understanding is essential for many applications and an integral part of self-driving cars. 
Particularly, fine-grained understanding provided by semantic segmentation is necessary to distinguish drivable and non-drivable surfaces and to reason about functional properties, like parking areas and sidewalks.
Currently, such understanding, represented in so-called high definition maps, is mainly generated in advance using surveying vehicles. 
However, self-driving cars should also be able to drive in unmapped areas and adapt their behavior if there are changes in the environment.

Most self-driving cars currently use multiple different sensors to perceive the environment. Complementary sensor modalities enable to cope with deficits or failures of particular sensors.
Besides cameras, light detection and ranging (LiDAR) sensors are often used as they provide precise distance measurements that are not affected by lighting.

\begin{table*}[t]
\centering
\small{
\begin{tabular}{lcccccc}
\toprule
 & \#scans$^1$ &\#points$^2$ & \#classes$^3$ & sensor & annotation & sequential  \\ 
\midrule
\textbf{\datasetname (Ours)} 	   			& \textbf{23201/20351} 	 & \textbf{4549} & \textbf{25} (28) & Velodyne HDL-64E	& point-wise & \cmark \\
Oakland3d \cite{munoz2009cvpr}		& 17 		& 1.6    & 5 (44) & SICK LMS 	& point-wise & \xmark   \\
Freiburg \cite{steder2010icra,behley2012icra}&	77	& 1.1	 & 4 (11) & SICK LMS	& point-wise & \xmark    \\
Wachtberg  \cite{behley2012icra}	& 5		& 0.4	 & 5 (5) & Velodyne HDL-64E & point-wise & \xmark    \\
Semantic3d \cite{hackel2017isprs} 	& 15/15 	& 4009   & 8 (8) & Terrestrial Laser Scanner & point-wise & \xmark  \\ 
Paris-Lille-3D \cite{roynard2018ijrr}	& 3 		& 143 	 & 9 (50) & Velodyne HDL-32E & point-wise & \xmark    \\
Zhang et al. \cite{zhang2015icra} 	& 140/112    &   32     & 10 (10) & Velodyne HDL-64E & point-wise & \xmark   \\
\midrule
KITTI \cite{geiger2012cvpr} 		& 7481/7518 	& 1799	 & 3  & Velodyne HDL-64E & bounding box & \xmark   \\
\bottomrule
\vspace{-0.5cm}
\label{tab:available_datasets}
\end{tabular}}
\caption{Overview of other point cloud datasets with semantic annotations. Ours is by far the largest dataset with sequential information. $^1$Number of scans for train and test set, $^2$Number of points is given in millions, $^3$Number of classes used for evaluation and number of classes annotated in brackets.}
\label{tab:datasets}
\vspace{-0.2cm}
\end{table*}

Publicly available datasets and benchmarks are crucial for empirical evaluation of research. They mainly fulfill three purposes:
(i)~they provide a basis to measure progress, since they allow to provide results that are reproducible and comparable,
(ii)~they uncover shortcomings of the current state of the art and therefore pave the way for novel approaches and research directions, and 
(iii)~they make it possible to develop approaches without the need to first painstakingly collect and label data. 
While multiple large datasets for \emph{image-based} semantic segmentation exist \cite{cordts2016cvpr, neuhold2017iccv}, publicly available datasets with point-wise annotation of three-dimensional point clouds are still comparably small, as shown in \reftab{tab:datasets}.

To close this gap we propose \emph{\datasetname}, a large dataset showing unprecedented detail in point-wise annotation with $28$ classes, which is suited for various tasks. In this paper, we mainly focus on \emph{laser-based} semantic segmentation, but also semantic scene completion. 
The dataset is distinct from other laser datasets as we provide accurate scan-wise annotations of sequences.
Overall, we annotated all $22$ sequences of the odometry benchmark of the \emph{KITTI Vision Benchmark} \cite{geiger2012cvpr} consisting of over $43\,000$  scans.
Moreover, we labeled the complete horizontal $360^\circ$ field-of-view of the rotating laser sensor.
Figure 1 shows example scenes from the provided dataset.
In summary, our main contributions are:
\begin{itemize}
 \itemsep0.1em
 \item We present a point-wise annotated dataset of point cloud sequences with an unprecedented number of classes and unseen level-of-detail for each scan.
 \item We furthermore provide an evaluation of state-of-the-art methods for semantic segmentation of point clouds. 
 \item We investigate the usage of sequence information  for semantic segmentation using multiple scans.
 \item Based on the annotation of sequences of a moving car, we furthermore introduce a real-world dataset for semantic scene completion and provide baseline results.
 \item Together with a benchmark website, the point cloud labeling tool is also publicly available, enabling other researchers to generate other labeled datasets in future.
\end{itemize}

This large dataset will stimulate the development of novel algorithms, make it possible to investigate new research directions, and puts evaluation and comparison of these novel algorithms on a more solid ground.

\section{Related Work}

The progress of computer vision has always been driven by benchmarks and datasets \cite{torralba2011cvpr}, but the availability of especially large-scale datasets, such as \emph{ImageNet}~\cite{deng2009cvpr}, was even a crucial prerequisite for the advent of deep learning.

More task-specific datasets geared towards self-driving cars were also proposed.
Notable is here the \emph{KITTI Vision Benchmark} \cite{geiger2012cvpr} since it showed that off-the-shelf solutions are not always suitable for autonomous driving.
The \emph{Cityscapes} dataset \cite{cordts2016cvpr} is the first dataset for self-driving car applications that provides a considerable amount of pixel-wise labeled images suitable for deep learning.
The \emph{Mapillary Vistas} dataset \cite{neuhold2017iccv} surpasses the amount and diversity of labeled data compared to \emph{Cityscapes}.

Also in point cloud-based interpretation, \eg, semantic segmentation, \mbox{RGB-D} based datasets enabled tremendous progress.
\emph{ShapeNet} \cite{shapenet2015tr} is especially noteworthy for point clouds showing a single object, but such data is not directly transferable to other domains.  
Specifically, LiDAR sensors usually do not cover objects as densely as an \mbox{RGB-D} sensor due to their lower angular resolution, in particular in vertical direction.

For indoor environments, there are several datasets \cite{silberman2012eccv, ros2016cvpr, hua2016threedv, armeni2017arxiv, dai2017cvpr, mcCormac2017iccv, li2018bmvc, dai2018cvpr} available, which are mainly recorded using RGB-D cameras or synthetically generated.
However, such data shows very different characteristics compared to outdoor environments, which is also caused by the size of the environment, since point clouds captured indoors tend to be much denser due to the range at which objects are scanned.
Furthermore, the sensors have different properties regarding sparsity and accuracy.
While laser sensors are more precise than RGB-D sensors, they usually only capture a sparse point cloud compared to the latter. 

For outdoor environments, datasets were recently proposed that are recorded with a terrestrial laser scanner (TLS), like the \emph{Semantic3d} dataset \cite{hackel2017isprs}, or using automotive LiDARs, like the \emph{Paris-Lille-3D} dataset \cite{roynard2018ijrr}.
However, the \emph{Paris-Lille-3D} provides only the aggregated scans with point-wise annotations for $50$ classes from which $9$ are selected for evaluation.
Another recently used large dataset for autonomous driving \cite{wang2018cvpr-dpcc}, but with fewer classes, is not publicly available.

The \emph{Virtual KITTI} dataset \cite{gaidon2016cvpr} provides synthetically generated sequential images with depth information and dense pixel-wise annotation.
The depth information can also be used to generate point clouds.
However, these point clouds do not show the same characteristics as a real rotating LiDAR, including defects like reflections and outliers.

In contrast to these datasets, our dataset combines a large amount of labeled points, a large variety of classes, and sequential scans generated by a commonly employed sensor used in autonomous driving, which is distinct from all publicly available datasets, also shown in \reftab{tab:datasets}.

\section{The \datasetname Dataset}

Our dataset is based on the odometry dataset of the KITTI Vision Benchmark~\cite{geiger2012cvpr} showing inner city traffic, residential areas, but also highway scenes and countryside roads around Karlsruhe, Germany.
The original odometry dataset consists of $22$ sequences, splitting sequences $00$~to~$10$ as training set, and $11$~to~$21$ as test set. 
For consistency with the original benchmark, we adopt the same division for our training and test set.
Moreover, we do not interfere with the original odometry benchmark by providing labels only for the training data.
Overall, we provide $23\,201$ full 3D scans for training and $20\,351$ for testing, which makes it by a wide margin the largest dataset publicly available.

We decided to use the KITTI dataset as a basis for our labeling effort, since it allowed us to exploit one of the largest available collections of raw point cloud data captured with a car.
We furthermore expect that there are also potential synergies between our annotations and the existing benchmarks and this will enable the investigation and evaluation of additional research directions, such as the usage of semantics for laser-based odometry estimation.

\begin{figure}
\centering
\includegraphics[width=0.78\linewidth]{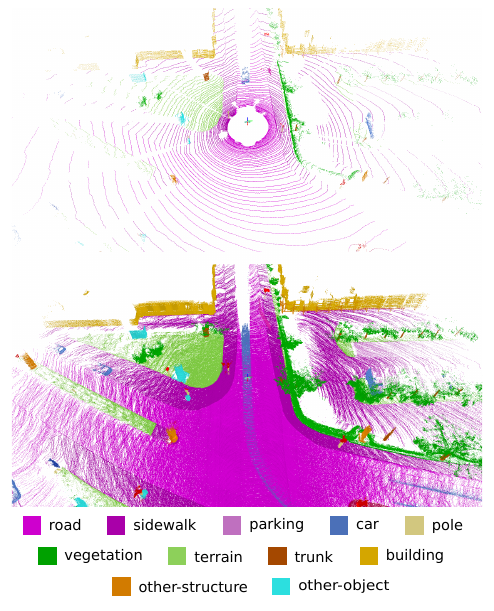}
\caption{Single scan (top) and multiple superimposed scans with labels (bottom).
Also shown is a moving car in the center of the image resulting in a trace of points.}
\label{fig:single_vs_aggregated}
\vspace{-0.45cm}
\end{figure}

Compared to other datasets (cf.\,\reftab{tab:datasets}), we provide labels for sequential point clouds generated with a commonly used automotive LiDAR, \ie, the Velodyne HDL-64E.
Other publicly available datasets, like \emph{Paris-Lille-3D} \cite{roynard2018ijrr}  or \emph{Wachtberg} \cite{behley2012icra}, also use such sensors, but only provide the aggregated point cloud of the whole acquired sequence or some individual scans of the whole sequence, respectively.
Since we provide the individual scans of the whole sequence, one can also investigate how aggregating multiple consecutive scans influences the performance of the semantic segmentation and use the information to recognize moving objects.

We annotated $28$ classes, where we ensured a large overlap of classes with the \emph{Mapillary Vistas} dataset \cite{neuhold2017iccv} and \emph{Cityscapes} dataset \cite{cordts2016cvpr} and made modifications where necessary to account for the sparsity and vertical field-of-view.
More specifically, we do not distinguish between persons riding a vehicle and the vehicle, but label the vehicle and the person as either \emph{bicyclist} or \emph{motorcyclist}.

\begin{figure*}[ht]
\centering
\includegraphics[width=0.95\textwidth]{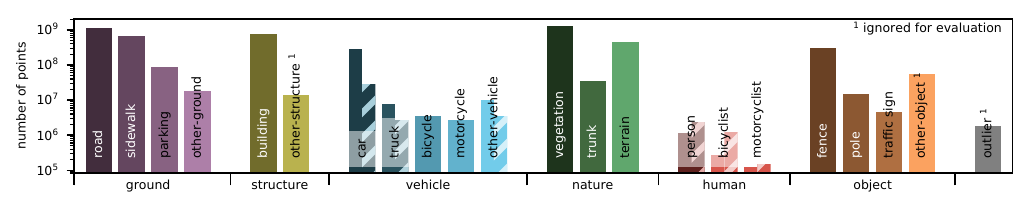}
\caption{Label distribution. The number of labeled points per class and the root categories for the classes are shown. For movable classes, we also show the number of points on non-moving (solid bars) and moving objects (hatched bars).}
\label{fig:label_distribution}
\vspace{-0.1cm}
\end{figure*}

We furthermore distinguished between moving and non-moving vehicles and humans, \ie, vehicles or humans gets the corresponding moving class if they moved in some scan while observing them, as shown in the lower part of \reffig{fig:single_vs_aggregated}.
All annotated classes are listed in \reffig{fig:label_distribution} and a more detailed discussion and definition of the different classes can be found in the supplementary material.
In summary, we have $28$ classes, where $6$ classes are assigned the attribute moving or non-moving, and one \emph{outlier} class is included for erroneous laser measurements caused by reflections or other effects.

The dataset is publicly available through a benchmark website and we provide only the training set with ground truth labels and perform the test set evaluation online.
We furthermore will also limit the number of possible test set evaluations to prevent overfitting to the test set~\cite{torralba2011cvpr}.

\subsection{Labeling Process}

To make the labeling of point cloud sequences practical, we superimpose multiple scans above each other, which conversely allows us to label multiple scans consistently.
To this end, we first register and loop close the sequences using an off-the-shelf laser-based SLAM system~\cite{behley2018rss}. 
This step is needed as the provided information of the inertial navigation system (INS) often results in map inconsistencies, \ie, streets that are revisited after some time  have different height.
For three sequences, we had to manually add loop closure constraints to get correctly loop closed trajectories, since this is essential to get consistent point clouds for annotation.
The loop closed poses allow us to load all overlapping point clouds for specific locations and visualize them together, as depicted in \reffig{fig:single_vs_aggregated}.

We subdivide the sequence of point clouds into tiles of $100$\,m by $100$\,m.
For each tile, we only load scans overlapping with the tile. 
This enables us to label all scans consistently even when we encounter temporally distant loop closures.
To ensure consistency for scans overlapping with more than one tile, we show all points inside each tile and a small boundary overlapping with neighboring tiles.
Thus, it is possible to continue labels from a neighboring tile.

Following best practices, we compiled a labeling instruction and provided instructional videos on how to label certain objects, such as cars and bicycles standing near a wall.
Compared to image-based annotation, the annotation process with point clouds is more complex, since the annotator often needs to change the viewpoint.
An annotator needs on average $4.5$ hours per tile, when labeling residential areas corresponding to the most complex encountered scenery, and needs on average $1.5$ hours for labeling a highway tile.

We explicitly did not use bounding boxes or other available annotations for the KITTI dataset, since we want to ensure that the labeling is consistent and the point-wise labels should only contain the object itself. % , \ie, we labeled ground beneath cars as road surfaces.

We provided regular feedback to the annotators to improve the quality and accuracy of labels.
Nevertheless, a single annotator also verified the labels in a second pass, \ie, corrected inconsistencies and added missing labels.
In summary, the whole dataset comprises $518$ tiles and over $1\,400$ hours of labeling effort have been invested with additional $10-60$ minutes verification and correction per tile, resulting in a total of over $1\,700$ hours.

\subsection{Dataset Statistics}

\reffig{fig:label_distribution} shows the distribution of the different classes, where we also included the root categories as labels on the x-axis.
The ground classes, \emph{road}, \emph{sidewalk}, \emph{building}, \emph{vegetation}, and \emph{terrain} are the most frequent classes.
The class  \emph{motorcyclist} only occurs rarely, but still more than $100\,000$ points are annotated.

The unbalanced count of classes is common for datasets captured in natural environments and some classes will be always under-represented, since they do not occur that often.
Thus, an unbalanced class distribution is part of the problem that an approach has to master.
Overall, the distribution and relative differences between the classes is quite similar in  other datasets, \eg \emph{Cityscapes} \cite{cordts2016cvpr}.

\section{Evaluation of Semantic Segmentation}
\label{sec:SemanticSegmentation}
In this section, we provide the evaluation of several state-of-the-art methods for semantic segmentation of a single scan. 
We also provide experiments exploiting information provided by sequences of multiple scans.

\subsection{Single Scan Experiments}

\paragraph{Task and Metrics.}
In semantic segmentation of point clouds, we want to infer the label of each three-dimensional point.
Therefore, the input to all evaluated methods is a list of coordinates of the three-dimensional points along with their remission, \ie, the strength of the  reflected laser beam which depends on the properties of the surface that was hit.
Each method should then output a label for each point of a scan, \ie, one full turn of the rotating LiDAR sensor.

To assess the labeling performance, we rely on the commonly applied mean Jaccard Index or mean intersection-over-union~(mIoU) metric~\cite{everingham2015ijcv} over
all classes, given by 
\begin{align}
&\frac{1}{C}~\sum_{c=1}^{C}\frac{\text{TP}_c}{\text{TP}_c + \text{FP}_c + \text{FN}_c}, \label{eq:miou}
\end{align}
where $\text{TP}_c$, $\text{FP}_c$, and $\text{FN}_c$ correspond to the number of true positive, false positive, and false negative predictions for class $c$, and $C$ is the number of classes.

As the classes \emph{other-structure} and \emph{other-object} have either only a few points and are otherwise too diverse with a high intra-class variation, we decided to not include these classes in the evaluation.
Thus, we use $25$ instead of $28$ classes, ignoring \emph{outlier}, \emph{other-structure}, and \emph{other-object} during training and inference.

Furthermore, we cannot expect to distinguish moving from non-moving objects with a single scan, since this Velodyne LiDAR cannot measure velocities like radars exploiting the Doppler effect.
We therefore combine the moving classes with the corresponding non-moving class resulting in a total number of $19$ classes for training and evaluation.

%% https://www.merriam-webster.com/dictionary/state%20of%20the%20art
\paragraph{State of the Art.}

Semantic segmentation or point-wise classification of point clouds is a long-standing topic \cite{anguelov2005cvpr}, which was traditionally solved using a feature extractor, such as Spin Images \cite{johnson1999pami}, in combination with a traditional classifier, like support vector machines \cite{agrawal2009icra} or even semantic hashing \cite{behley2010iros}.
Many approaches used Conditional Random Fields (CRF) to enforce label consistency of neighboring points~\cite{triebel2006icra, munoz20083dpvt, munoz2009cvpr, munoz2009icra, xiong2011icra}.

With the advent of deep learning approaches in image-based classification, the whole pipeline of feature extraction and classification has been replaced by end-to-end deep neural networks. 
Voxel-based methods transforming the point cloud into a voxel-grid and then applying convolutional neural networks (CNN) with 3D convolutions for object classification \cite{maturana2015iros} and semantic segmentation \cite{huang2016icpr} were among the first investigated models, since they allowed to exploit architectures and insights known for images.

To overcome the limitations of the voxel-based representation, such as the exploding memory consumption when the resolution of the voxel grid increases, more recent approaches either upsample voxel-predictions~\cite{tchapmi2017threedv} using a CRF or use different representations, like more efficient spatial subdivisions~
\cite{klokov2017iccv,riegler2017cvpr,zeng2017arxiv,wang2018arxiv,graham2018cvpr}, rendered 2D image views~\cite{boulch2017cg}, graphs \cite{landrieu2018cvpr,te2018arxiv}, splats~\cite{su2018cvpr}, or even directly the
 points~\cite{qi2017nips, qi2017cvpr, hua2018cvpr, groh2018accv,rethage2018eccv,jiang2018arxiv,engelmann2018arxiv}.

\input{script/table_result_seg.tex}

\paragraph{Baseline approaches.}

We provide the results of six state-of-the-art architectures for the semantic segmentation of
point clouds in our dataset: PointNet~\cite{qi2017cvpr}, PointNet++~\cite{qi2017nips},  Tangent Convolutions~\cite{tatarchenko2018cvpr}, SPLATNet~\cite{su2018cvpr},
\ifspgraph Superpoint Graph~\cite{landrieu2018cvpr}, \fi
and SqueezeSeg~(V1 and V2)~\cite{wu2018icra-scnn,wu2019icra}. 
Furthermore, we investigate two extensions of SqueezeSeg: DarkNet21Seg and DarkNet53Seg.

PointNet \cite{qi2017cvpr} and PointNet++~\cite{qi2017nips} use the raw un-ordered point cloud data as input.
Core of these approaches is max pooling to get an order-invariant operator that works surprisingly well for semantic segmentation of shapes and several other benchmarks. 
Due to this nature, however, PointNet fails to capture the spatial relationships between the features. 
To alleviate this, PointNet++~\cite{qi2017nips} applies individual PointNets to local neighborhoods and uses a hierarchical approach to combine their outputs.
This enables it to build complex hierarchical features that capture both local fine-grained and global contextual information.

Tangent Convolutions~\cite{tatarchenko2018cvpr} also handles unstructured point clouds by applying convolutional neural networks directly on surfaces.
This is achieved by assuming that the data is sampled from smooth surfaces and defining a tangent convolution as a convolution applied to the projection of the local surface at each point into the tangent plane.

SPLATNet~\cite{su2018cvpr} takes an approach that is similar to the aforementioned voxelization methods and represents the point clouds in a high-dimensional sparse lattice.
As with voxel-based methods, this scales poorly both in computation and in memory cost and therefore they exploit the sparsity of this representation by using bilateral convolutions~\cite{jampani2016cvpr}, which only operates on occupied lattice parts.

\ifspgraph
Similarly to PointNet, Superpoint Graph~\cite{landrieu2018cvpr}, captures the local relationships by summarizing geometrically homogeneous groups of points into superpoints, which are later embedded by local PointNets.
The result is a superpoint graph representation that is more compact and rich than the original point cloud exploiting contextual relationships between the superpoints. 
\fi

SqueezeSeg~\cite{wu2018icra-scnn,wu2019icra} also discretizes the point cloud in a way that makes it possible to apply 2D convolutions to the point cloud data exploiting the sensor geometry of a rotating LiDAR.
In the case of a rotating LiDAR, all points of a single turn can be projected to an image by using a spherical projection. 
A fully convolutional neural network is applied and then finally filtered with a CRF to smooth the results.
Due to the promising results of SqueezeSeg and the fast training, we investigated how the labeling performance is affected by the number of model parameters. 
To this end, we used a different backbone based on the Darknet architecture~\cite{redmon2018arxiv} with $21$ and $53$ layers, and $25$ and $50$ million parameters respectively. 
We furthermore eliminated the vertical downsampling used in the architecture.

We modified the available implementations such that the methods could be trained and evaluated on our large-scale dataset. 
Note that most of these approaches have so far only been evaluated on shape \cite{shapenet2015tr} or RGB-D indoor datasets \cite{silberman2012eccv}.
However, some of the approaches~\cite{qi2017cvpr,qi2017nips} were only possible to run with considerable downsampling to $50\,000$ points due to memory limitations.

\paragraph{Results and Discussion.}

\reftab{tab:results_segmentation} shows the results of our baseline experiments for various approaches using either directly the point cloud information~\cite{qi2017cvpr,qi2017nips,su2018cvpr,tatarchenko2018cvpr,landrieu2018cvpr} or a projection of the point cloud~\cite{wu2018icra-scnn}.
The results show that the current state of the art for point cloud semantic segmentation falls short for the size and complexity of our dataset. 

We believe that this is mainly caused by the limited capacity of the used architectures (see \reftab{tab:approach_stats}), because the number of parameters of these approaches is much lower than the number of parameters used in leading image-based semantic segmentation networks.
As mentioned above, we added DarkNet21Seg and DarkNet53Seg to test this hypothesis and the results show that this simple modification improves the accuracy from $29.5\,\%$ for SqueezeSeg to $47.4\,\%$ for DarkNet21Seg and to $49.9\,\%$ for DarkNet53Seg.

Another reason is that the point clouds generated by LiDAR are relatively sparse, especially as the distance to the sensor increases. 
This is partially solved in SqueezeSeg, which exploits the way the rotating scanner captures the data to generate a dense range image, where each pixel corresponds roughly to a point in the scan. 

\begin{figure}[t]
\centering
\vspace{-0.25cm}
\includegraphics[width=\linewidth]{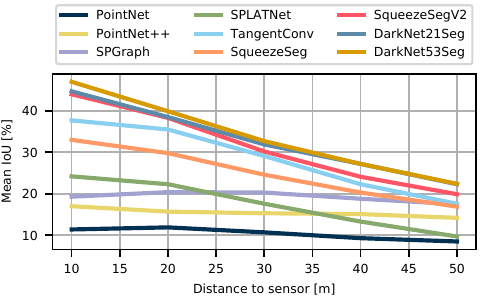}
\caption{IoU vs. distance to the sensor.}
\vspace{-0.4cm}
\label{fig:iou-vs-distance}
\end{figure}

These effects are further analyzed in \reffig{fig:iou-vs-distance}, where the mIoU is plotted \wrt the distance to the sensor. 
It shows that results of all approaches get worse with increasing distance.
This further confirms our hypothesis that the sparsity is the main reason for worse results at large distances.
However, the results also show that some methods, like SPGraph, are less affected by the distance-dependent sparsity and this might be a promising direction for future research to combine the strength of both paradigms.

Especially classes with few examples, like motorcyclists and trucks, seem to be more difficult for all approaches. 
But also classes with only a small number of points in a single point cloud, like bicycles and poles, are hard classes.

\begin{table}
 \centering
 \vspace{-0.1cm}
\footnotesize{
 \begin{tabular}{lccc}
 \toprule
  Approach & num. parameters & train time & inference time \\
  & (million)& $\left(\frac{\text{GPU hours}}{\text{epoch}}\right)$ & $\left(\frac{\text{seconds}}{\text{point cloud}}\right)$ \\
  \midrule 
  PointNet & $3$ & $4$ & $0.5$ \\
  PointNet++ & $6$ & $16$ & $5.9$ \\
  \ifspgraph
  SPGraph  & $0.25$ & $6$ & $5.2$ \\
  \fi
  TangentConv & $0.4$ & $6$ & $3.0$ \\
  SPLATNet & $0.8$ & $8$ & $1.0$ \\
  SqueezeSeg & $1$ & $0.5$ & $0.015$ \\
  SqueezeSegV2 & $1$ & $0.6$ & $0.02$ \\
  DarkNet21Seg & $25$ & $2$ & $0.055$ \\
  DarkNet53Seg & $50$ & $3$ & $0.1$ \\
  \bottomrule
 \end{tabular}}
 \caption{Approach statistics.}
 \label{tab:approach_stats}
 \vspace{-0.25cm}
\end{table}

Finally, the best performing approach (DarkNet53Seg) with $49.9\%$ mIoU is still far from achieving results that are on par with image-based approaches, \eg, $80\%$ on the \emph{Cityscapes} benchmark \cite{cordts2016cvpr}.

\subsection{Multiple Scan Experiments}

\paragraph{Task and Metrics.}

In this task, we allow methods to exploit information from a sequence of multiple past scans to improve the segmentation of the current scan. 
We furthermore want the methods to distinguish moving and non-moving classes, \ie, all $25$ classes must be predicted, since this information should be visible in the temporal information of multiple past scans.
The evaluation metric for this task is still the same as in the single scan case, \ie, we evaluate the mean IoU of the current scan no matter how many past scans were used to compute the results. % but in all $25$ classes.

\paragraph{Baselines.}

We exploit the sequential information by combining $5$ scans into a single, large point cloud, \ie, the current scan at timestamp $t$ and the $4$ scans before at timestamps \mbox{$t-1, \dots, t-4$}.
We evaluate DarkNet53Seg and TangentConv, since these approaches can deal with a larger number of points without downsampling of the point clouds and could still be trained in a reasonable amount of time.

\paragraph{Results and Discussion.}

\reftab{tab:results_multiscan} shows the per-class results for the movable classes and the mean IoU (mIoU) over all classes.
For each method, we show in the upper part of the row the IoU for non-moving (unshaded) and in the lower part of the row  the IoU for moving objects (shaded). 
The performance of the remaining static classes is similar to the single scan results and we refer to the supplement for a table containing all classes.

The general trend that the projective methods perform better than the point-based methods is still apparent, which can be also attributed to the larger amount of parameters as in the single scan case.
Both approaches show difficulties in separating moving and non-moving objects, which might be caused by our design decision to aggregate multiple scans into a single large point cloud.
The results show that especially bicyclist and motorcyclist never get correctly assigned the non-moving class, which is most likely a consequence from the generally sparser object point clouds.

We expect that new approaches could explicitly exploit the sequential information by using multiple input streams to the architecture  or even recurrent neural networks to account for the temporal information, which again might open a new line of research.

\input{script/table_result_multiscan.tex}

%%%%%%%%%%%%%%%%%%%%%%%%%%%%%%%%%%%%%%%%%%%%%%%%%%%%%%%%%%%%%%%%%%%%%%%%%%%%%%%%%%%%%%%%%%%%%%%%%%%%%%%%%

\section{Evaluation of Semantic Scene Completion}

After leveraging a sequence of past scans for semantic point cloud segmentation, we now show a scenario that makes use of future scans.
Due to its sequential nature, our dataset provides the unique opportunity to be extended for the task of 3D semantic scene completion.
Note that this is the first real world outdoor benchmark for this task. 
Existing point cloud datasets cannot be used to address this task, as they do not allow for aggregating labeled point clouds that are sufficiently dense in both space and time.

In semantic scene completion, one fundamental problem is to obtain ground truth labels for real world datasets. 
In case of NYUv2~\cite{silberman2012eccv}, CAD models were fit into the scene~\cite{rock2015cvpr} using an RGB-D image captured by a Kinect sensor.
New approaches  often resort to prove their effectiveness on the larger, but synthetic SUNCG dataset~\cite{song2017cvpr}. 
However, a dataset combining the scale of a synthetic dataset and usage of real-world data is still missing.

In the case of our proposed dataset, the car carrying the LiDAR moves past 3D objects in the scene and thereby records their backsides, which are hidden in the initial scan due to self-occlusion.
This is exactly the information needed for semantic scene completion as it contains the full 3D geometry of all objects while their semantics are provided by our dense annotations.

\paragraph{Dataset Generation. } 
By superimposing an exhaustive number of future laser scans in a predefined region in front of the car, we can generate pairs of inputs and targets that correspond to the task of semantic scene completion. 
As proposed by Song \etal~\cite{song2017cvpr}, our dataset for the scene completion task is a voxelized representation of the 3D scene.

We select a volume of $51.2\,$m ahead of the car, $25.6\,$m to every side and $6.4\,$m in height with a voxel resolution of $0.2\,$m, which results in a volume of $256 \times 256 \times 32$ voxels to predict. 
We assign a single label to every voxel based on the majority vote over all labeled points inside a voxel.
Voxels that do not contain any points are labeled as \textit{empty}.

\begin{figure*}
  \centering
  \includegraphics[width=0.9\textwidth]{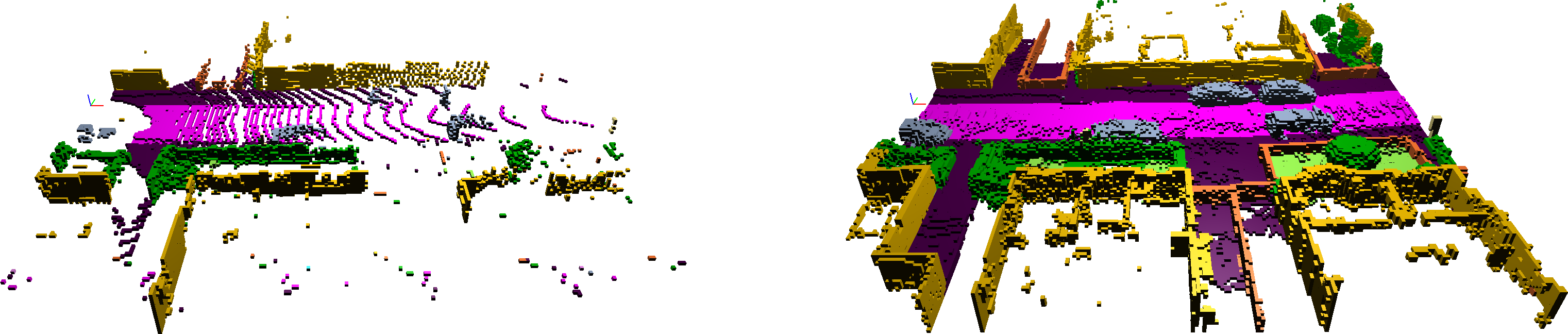}
  \caption{Left: Visualization of the incomplete input for the semantic scene completion benchmark. Note that we show the labels only for better visualization, but the real input is a single raw voxel grid without any labels. Right: Corresponding target output representing the completed and fully labeled 3D scene.}
  \vspace{-0.2cm}
  \label{fig:ssc_qual_results}
\end{figure*}

To compute which voxels belong to the occluded space, we check for every pose of the car which voxels are visible to the sensor by tracing a ray. 
Some of the voxels, \eg those inside objects or behind walls are never visible, so we ignore them during training and evaluation.

Overall, we extracted  $19\,130$ pairs of input and target voxel grids for training, $815$ for validation and $3\,992$ for testing.
For the test set, we only provide the unlabeled input voxel grid and withhold the target voxel grids. \reffig{fig:ssc_qual_results} shows an example of an input and target pair.

\paragraph{Task and Metrics.}
In semantic scene completion, we are interested in predicting the complete scene inside a certain volume from a single initial scan.
More specifically, we use as input a voxel grid, where each voxel is marked as empty or occupied, depending on whether or not it contains a laser measurement.
For semantic scene completion, one needs to predict whether a voxel is occupied and its semantic label in the completed scene.

For evaluation, we follow the evaluation protocol of Song \etal~\cite{song2017cvpr} and compute the IoU for the task of scene completion, which only classifies a voxel as being occupied or empty, \ie, ignoring the semantic label, as well as mIoU~\eqref{eq:miou} for the task of semantic scene completion over the same $19$ classes that were used for the single scan semantic segmentation task (see Section~\ref{sec:SemanticSegmentation}).

\paragraph{State of the Art. }
Early approaches addressed the task of scene completion either without predicting semantics~\cite{firman2016cvpr}, thereby not providing a holistic understanding of the scene, or by trying to fit a fixed number of mesh models to the scene geometry~\cite{geiger2015gcpr}, which limits the expressiveness of the approach.

Song \etal~\cite{song2017cvpr} were the first to address the task of semantic scene completion in an end-to-end fashion. 
Their work spawned a lot of interest in the field yielding models that combine the usage of color and depth information~\cite{liu2018nips,garbade2019cvpr-ws} or address the problem of sparse 3D feature maps by introducing submanifold convolutions~\cite{zhang2018eccv} or increase the output resolution by deploying a multi-stage coarse to fine training scheme~\cite{dai2018cvpr}. 
Other works experimented with new encoder-decoder CNN architectures as well as improving the loss term by adding adversarial loss components~\cite{wang2018threedv}.% TODO-MG: Also add Chen et al (under review)

\paragraph{Baseline Approaches.}
We report the results of four semantic scene completion approaches.
In the first approach, we apply SSCNet~\cite{song2017cvpr} without the flipped TSDF as input feature. 
This has minimal impact on the performance, but significantly speeds up the training time due to faster preprocessing~\cite{garbade2019cvpr-ws}.
Then we use the Two Stream (TS3D) approach~\cite{garbade2019cvpr-ws}, which makes use of the additional information from the RGB image corresponding to the input laser scan. 
Therefore the RGB image is first processed by a 2D semantic segmentation network, using the approach DeepLab v2 (ResNet-101)~\cite{chen2018pami} trained on Cityscapes to generate a semantic segmentation. %TODO: Check whether the network was finetuned on Kitti Sem Seg benchmark and if yes, how this changed the output class distribution
The depth information from the single laser scan and the labels inferred from the RGB image are combined in an early fusion.
Furthermore, we modify the TS3D approach in two steps: 
First, by directly using labels from the best LiDAR-based semantic segmentation approach (DarkNet53Seg) and secondly, by exchanging the 3D-CNN backbone by SATNet~\cite{liu2018nips}.

\paragraph{Results and Discussion.}
\reftab{tab:ssc_baselines} shows the results of each of the baselines, whereas results for individual classes are reported in the supplement.
The TS3D network, incorporating 2D semantic segmentation of the RGB image, performs similar to SSCNet which only uses depth information.
However, the usage of the best semantic segmentation directly working on the point cloud  slightly outperforms SSCNet on semantic scene completion (TS3D + DarkNet53Seg). 
Note that the first three approaches are based on SSCNet's 3D-CNN architecture, which performs a 4 fold downsampling in a forward pass and thus renders them incapable of dealing with details of the scene.
In our final approach, we exchange the SSCNet-backbone of TS3D + DarkNet53Seg with SATNet~\cite{liu2018nips}, which is capable of dealing with the desired output resolution.
Due to memory limitations, we use random cropping during training. During inference, we divide each volume into six equal parts, perform scene completion on them individually and subsequently fuse them. This approach performs much better than the SSCNet based approaches.

Apart from dealing with the target resolution, a challenge for current models is the sparsity of the laser input signal in the far field as can be seen from \reffig{fig:ssc_qual_results}. To obtain a higher resolution input signal in the far field, approaches would have to exploit more efficiently information from high resolution RGB images provided along with each laser scan.

\begin{table}[t]
  \centering
\resizebox{\columnwidth}{!}{
\footnotesize{
\begin{tabular}{lcc}
\toprule
    & Completion & Semantic Scene \\
    &     (IoU)            & Completion (mIoU) \\
\midrule
SSCNet \cite{song2017cvpr}                   & 29.83 & 9.53 \\
TS3D \cite{garbade2019cvpr-ws}             & 29.81 & 9.54 \\
TS3D \cite{garbade2019cvpr-ws} + DarkNet53Seg & 24.99 & 10.19 \\ 
TS3D \cite{garbade2019cvpr-ws} + DarkNet53Seg + SATNet & 50.60 & 17.70 \\
\bottomrule
\end{tabular}}
}
 \caption{Semantic scene completion baselines.}
 \label{tab:ssc_baselines}
 \vspace{-0.4cm}
\end{table}

\section{Conclusion and Outlook}

In this work, we have presented a large-scale dataset showing unprecedented scale in point-wise annotation of point cloud sequences.
We provide a range of different baseline experiments for three tasks: (i) semantic segmentation using a single scan, (ii) semantic segmentation using multiple scans, and (iii) semantic scene completion.

In future work, we plan to provide also instance-level annotation over the whole sequence, \ie, we want to distinguish different objects in a scan, but also identify the same object over time.
This will enable to investigate temporal instance segmentation over sequences.
However, we also see potential for other new tasks based on our labeling effort, such as the evaluation of semantic SLAM.

{\small  
\paragraph{Acknowledgments} We thank all students that helped with annotating the data. 
The work has been funded by~the  Deutsche Forschungsgemeinschaft (DFG, German~Research~Foundation) 
under FOR 1505 Mapping on Demand,  \mbox{BE~5996/1-1}, \mbox{GA~1927/2-2}, and under Germany’s Excellence Strategy, EXC-2070 -- 390732324 (PhenoRob).

}
{\small
\bibliographystyle{ieee_fullname}
\bibliography{references}
}

\appendix

\section{Consistent Labels for LiDAR Sequences}

In this section, we explain the implementation of our point cloud labeling tool in more detail and the rationale behind our decision to subdivide the sequences spatially, but not temporally, for getting consistently labeled point cloud sequences.
The labeling tool itself was critical to provide the amount of scans with such fine-grained labels.

\begin{figure}[t]
\centering
\includegraphics[width=0.95\linewidth]{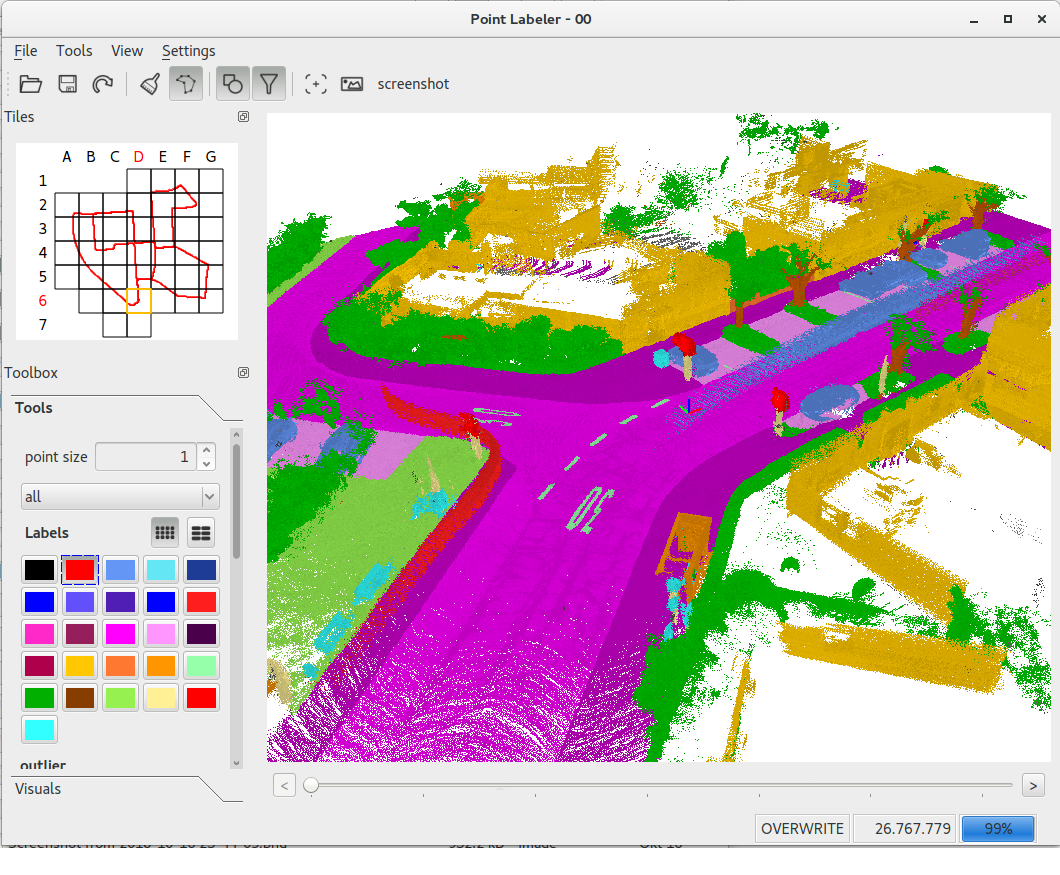}
\caption{Point cloud labeling tool. In the upper left corner the user sees the tile and the sensor's path indicated by the red trajectory. }
\label{fig:pca}
\end{figure}

In summary, we developed an OpenGL-based labeling tool, which exploits parallelization on the GPU.
The main challenge is the visualization of vast amounts of point data, but also processing these at the same time, while reaching responsiveness that allows the annotator to label interactively the aggregated point clouds.
\reffig{fig:pca} shows our point cloud annotation program visualizing an aggregated point cloud of over $20$ million points.
We provide a wide range of tools for annotation, like a brush, a polygon tool, and different filtering methods to hide selected labels.
Even with that many points, we are still able to maintain interactive labeling capabilities.
Changes to the label of the points inside the aggregated point cloud are reflected in the individual scans, which enables high consistency of the labels over time.

Since we are labeling each point, we are able to annotate objects, even with complex occlusions, more precisely than just using bounding volumes~\cite{xie2016cvpr}.
For instance, we ensured that ground points below a car are labeled accordingly, which was enabled by our filtering capabilities of the annotation tool.

To accelerate the search for points that must be labeled, we used a projective approach to assign labels.
To this end, we determine for each point the two-dimensional projection on the screen and then determine for the projection if the point is near to the clicked position (in case of the brush) or inside the selected polygon.
Therefore, annotators had to ensure that they did not choose a view that essentially destroyed previously assigned points.

Usually, an annotator performed the following cycle to annotate points: (1) mark points with a specific label and (2) filter points with that label.
Due to the filtering of already labeled points, one can resolve occlusions and furthermore ensure that the aforementioned projective labeling does not destroy already labeled points.

\paragraph{Tile-Based Labeling.} 
An important detail is the aforementioned spatial subdivision of the complete aggregated point cloud into tiles (also shown in the left upper part of \reffig{fig:pca}).
Initially, we simply rendered all scans in a range of timestamps, say $100-150$, and then moved on the next part, say $150-200$.
However, this leads quickly to inconsistencies in the labels, since scans from such parts still overlap and therefore must be relabeled to match labels from before.
Since we, furthermore, encounter loop closures with a considerable temporal distance, this overlap can even happen between parts of the sequences that are not temporally close, which even more complicated the task.

Thus, it quickly became apparent that such an additional effort to ensure consistent labels would lead to an unreasonable complicated annotations process and consequently to insufficient results.
Therefore, we decided to subdivide the sequence spatially into tiles, where each tile contains all points from scans overlapping with this tile.
Consistency at the boundaries between tiles was achieved by having a small overlap between the tiles, which enabled to consistently continue the labels from one tile into another neighboring tile. 

\paragraph{Moving Objects. }
We annotated all moving objects, i.e., car, truck, person, bicyclist, and motorcyclist, and each moving object is represented by a different class to distinguish it from its corresponding non-moving class.
In our case, we assigned an object the corresponding moving class when it moved at some point in time while observing it with the sensor.

Since moving objects will appear at different places when aggregating scans captured from different sensor locations, we had to take special care to annotate moving objects.
This is especially challenging, when multiple types of vehicles move on the same lane, like in most of the encountered highway scenes.
We annotated moving objects either by filtering ground points or by labeling each scan individually, which was often necessary to label points of tires of a car and bicycles or the feet of persons.
But scan-by-scan labeling was also necessary in aforementioned cases where multiple vehicles of different type drive on the same lane.
The labeling of moving objects often was the first step when annotating a tile, since this allowed the annotator to filter all moving points and then concentrate on the static parts of the environment.

\section{Basis of the Dataset}

The basis of our dataset is data from the KITTI Vision Benchmark~\cite{geiger2012cvpr}, which is still the largest collection of data also used in autonomous driving at the time of writing.
The KITTI dataset is the basis of many experimental evaluations in different contexts and was extended by novel tasks or additional data over time.
Thus, we decided to build upon this legacy and also enable synergies between our annotations and other parts and tasks of the KITTI Vision Benchmark.

We particularly decided to use the Odometry Benchmark to enable usage of the annotation data with this task.
We expect that exploiting semantic information in the odometry estimation is an interesting avenue for future research.
However, also other tasks of the KITTI Vision Benchmark might profit from our annotations and the pre-trained models we will publish on the dataset website.

Nevertheless, we hope that our effort and the availability of the point labeling tool will enable others to replicate our work on future publicly available datasets from an automotive LiDAR.

\begin{table*}
\centering
\begin{tabular}{p{0.3cm}lp{12cm}}
% \begin{tabular}{llp{8cm}c}
 \toprule
 cat. & class & definition \\
 \midrule
 %% the rows are ``lines`` of the table.
 \multirow{12}{*}{\begin{sideways}Ground-related\end{sideways}} 
 
 & road & Drivable areas where cars are allowed to drive on including service lanes, bike lanes, crossed areas on the street. Only the road surface is labeled excluding the curb. \\

 & sidewalk & Areas used mainly by pedestrians, bicycles, but not meant for driving with a car. This includes curbs and spaces where you are not allowed to drive faster than \SI{5}{\km~/~\hour}. Private driveways are also labeled as sidewalk. Here cars should also not drive with regular speeds (such as \num{30} or \SI{50}{\km~/~\hour}). \\

& parking & Areas meant explicitly for parking and that are clearly  separated from sidewalk and road by means of a small curb. If unclear then \textit{other-ground} or \textit{sidewalk} can be selected. Garages are labeled as \textit{building} and not as parking. \\

& other-ground & This label is chosen whenever a distinction between sidewalk and terrain is unclear. It includes (paved/plastered) traffic islands which are not meant for walking. Also the paved parts of a gas station are not meant for parking. \\

\midrule
\multirow{3}{*}{\begin{sideways}structures\end{sideways}} 

& building & The whole building including building walls, doors, windows, stairs, etc. Garages count as building. \\

& other-structure & This includes other vertical structures, like tunnel walls, bridge posts, scaffolding on a building from a construction site  or bus stops with a roof. \\

\midrule
\multirow{13}{*}{\begin{sideways}vehicle\end{sideways}} 

& car & Cars, jeeps, SUVs, vans with a continuous body shape (i.e. the driver cabin and cargo compartment are one) are included.  \\

& truck & Trucks, vans with a body that is separate from the driver cabin, pickup trucks, as well as their attached trailers.  \\

& bicycle & Bicycles without the cyclist or possibly other passengers. If the bicycle is driven by a person or a person stands nearby the vehicle, we label it as bicyclist.  \\

& motorcycle & Motorcycles, mopeds without the driver or other passengers. Includes also motorcycles covered by a cover. If the motorcycle is driven by a person or a person stands nearby the vehicle, we label it as motorcyclist.  \\

& other-vehicle & Caravans, Trailers and fallback category for vehicles not explicitly defined otherwise in the meta category \textit{vehicle}. Included are buses intended for \num{9}+ persons for public or long-distance transport. This further includes all vehicles moving on rails, e.g., trams, trains.  \\

\midrule
\multirow{4}{*}{\begin{sideways}nature\end{sideways}}

 & vegetation & Vegetation are all bushes, shrubs, foliage, and other clearly identifiable vegetation. \\

 & trunk &  The tree trunk is labeled as \textit{trunk} separately from the treetop which gets the label \textit{vegetation}.  \\

 & terrain & Grass and all other types of horizontal spreading vegetation, including soil.  \\
 
\midrule
\multirow{6}{*}{\begin{sideways}human\end{sideways}} & person & Humans moving by their own legs, sitting, or any unusual pose, but not meant to drive a vehicle. \\

 & bicyclist & Humans driving a bicycle or standing in close range to a bicycle (within \textit{arm reach}). We do not distinguish between riders and bicyclist.
 \\
 
 & motorcyclist & Humans driving a motorcycle or standing in close range to a motorcycle (within \textit{arm reach}). \\
 
\midrule
\multirow{5}{*}{\begin{sideways}object\end{sideways}} & fence & Separators, like fences, small walls and crash barriers. \\

& pole & Lamp posts and the poles of traffic signs. \\

& traffic sign & Traffic sign excluding its mounting. Spurious points in a layer in front and behind the traffic sign are also labeled as traffic sign and not as outlier. \\

& other-object & Fallback category that includes advertising columns. \\

\midrule
\multirow{2}{*}{\begin{sideways}outlier\end{sideways}} & outlier & Outlier are caused by reflections or inaccuracies in the deskewing of scans, where it is unclear where the points came from. \\
  \bottomrule
  \end{tabular}
 \caption{Class definitions.}
 \label{tab:class_def}

\end{table*}

%%%%%%%%% BODY TEXT
\section{Class Definition}

In the process of labeling such large amounts of data, we had to decide which classes we want to be annotated at some point in time.
In general, we followed the class definitions and selection of the \emph{Mapillary Vistas} dataset \cite{neuhold2017iccv} and \emph{Cityscapes} \cite{cordts2016cvpr} dataset, but did some simplifications and adjustments for the data source used.

First, we do not explicitly consider a \emph{rider} class for persons riding a motorcycle or a bicycle, since the available point clouds do not provide the density for a single scan to distinguish the person riding a vehicle.
Furthermore, we get for such classes only moving examples and therefore cannot easily aggregate the point clouds to increase the fidelity of the point cloud and make it easier to distinguish the rider of a vehicle and the vehicle.

The classes \emph{other-structure}, \emph{other-vehicle}, and \emph{other-object} are fallback classes of their respective root category in unclear cases or missing classes, since this simplified the labeling process and might be used to distinguish these categories further in future.

Annotators often annotated some object or part of the scene and then hide the labeled points to avoid overwriting or removing the labels.
Thus, assigning the fallback class in ambiguous cases  or cases where a specific class was missing made it possible to simply hide that class to avoid overwriting it.
If we had instructed the annotators to label such parts as unlabeled, it would have caused problems to consistently label the point clouds.

We furthermore distinguished between moving and non-moving vehicles and humans, \ie, a vehicle or human gets the `moving' tag if it moved in some consecutive scans while being observed by the LiDAR sensor.

In summary, we annotated $28$ classes and all annotated classes with their respective definitions are listed in \reftab{tab:class_def} on the next page.

 %\newpage
\begin{table}
  \centering
 \resizebox{\columnwidth}{!}{
 \footnotesize{
   \setlength\tabcolsep{4.5pt}
  \begin{tabular}{llccccc}
  \toprule
    & Approach  & 
   \begin{sideways}scan size \end{sideways} & 
   \begin{sideways}projected \end{sideways} & 
   \begin{sideways}learning rate \end{sideways} & 
   \begin{sideways}epochs trained\end{sideways} &  
   \begin{sideways}converged\end{sideways} \\
   \midrule 
   \multirow{7}{*}{\begin{sideways}single scan\end{sideways}}
   & PointNet & \num{50000} & - & $3\cdot 10^{-4}\!\times\!0.9^\text{epoch}$ & \num{33} &\cmark \\
   & PointNet++ & \num{45000} & - & $3\cdot 10^{-3}\!\times\!0.9^\text{epoch}$ & \num{25} & \cmark  \\
   & TangentConv & \num{120000} & - & $1\cdot 10^{-4}$ & \num{10} &\cmark  \\
   & SPLATNet & \num{50000} & - & $1\cdot 10^{-3}$ & \num{20} & -\\
   & SqueezeSeg & \num{64}$\,\times\,$\num{2048} & \cmark &  $1\cdot 10^{-2}\!\times\!0.99^\text{epoch}$ & $200$ & \cmark \\
   & DarkNet21Seg & \num{64}$\,\times\,$\num{2048} &\cmark  & $1\cdot 10^{-3}\!\times\!0.99^\text{epoch}$ & $40$ & \cmark  \\
   & DarkNet53Seg & \num{64}$\,\times\,$\num{2048} & \cmark  & $1\cdot 10^{-3}\!\times\!0.99^\text{epoch}$ & $120$ & \cmark  \\
   \midrule
   \multirow{2}{*}{\begin{sideways}\makecell{multi \\ scan}\end{sideways}}
   & TangentConv & \num{500000} & - & & $\num{5}^\ast$ & \cmark  \\
   & DarkNet53Seg64 & \num{64}$\,\times\,$\num{2048} & \cmark  & $1\cdot 10^{-3}\!\times\!0.99^\text{epoch}$  & $\num{40}^\ast$ & \cmark  \\
   \bottomrule
  \end{tabular}}
  }
  \caption{Approach statistics. $\ast$ in number of epochs means that it was started from the pretrained
  weights of the single scan version.}
  \label{tab:approach_stats}
 \end{table} 

%
% \clearpage 
% \onecolumn

\begin{figure*}
\centering
\includegraphics[width=\textwidth]{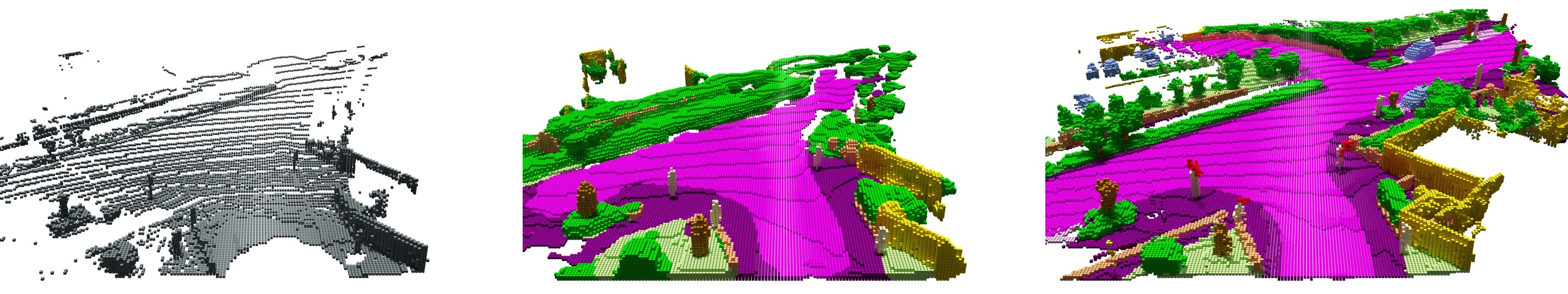} 
\caption{Qualitative results for the semantic scene completion approach TS3D + DarkNet53Seg + SATNet. Left: Input volume. Middle: Network prediction. Right: Ground truth. Due to memory limitations the inference has to be done in six steps on overlapping subvolumes. The subvolumes are consequently fused to obtain the final result.}
\label{fig:ssc_results}
\end{figure*}

\newlength{\resultsize}
\setlength{\resultsize}{0.8\textwidth} 
\begin{figure*}[th]
\centering
PointNet~\cite{qi2017cvpr}\\ \includegraphics[width=\resultsize]{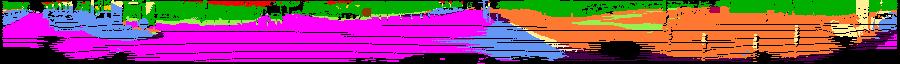} \\
\ifspgraph
SPGraph~\cite{landrieu2018cvpr}\\ \includegraphics[width=\resultsize]{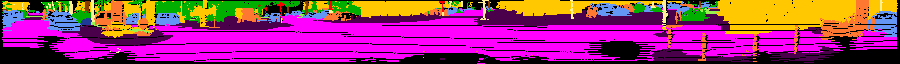}\\
\fi
SPLATNet~\cite{su2018cvpr}\\ \includegraphics[width=\resultsize]{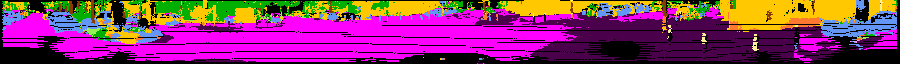}\\
PointNet++~\cite{qi2017nips}\\ \includegraphics[width=\resultsize]{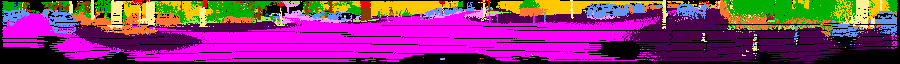} \\
SqueezeSeg~\cite{wu2018icra-scnn}\\ \includegraphics[width=\resultsize]{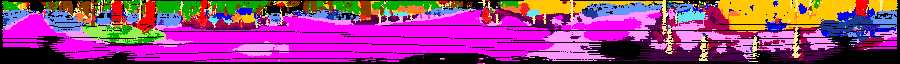}\\
TangentConv~\cite{tatarchenko2018cvpr}\\ \includegraphics[width=\resultsize]{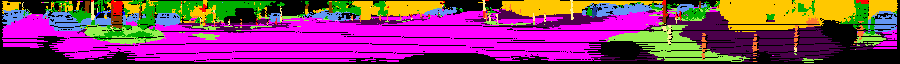}\\
Darknet21Seg\\ \includegraphics[width=\resultsize]{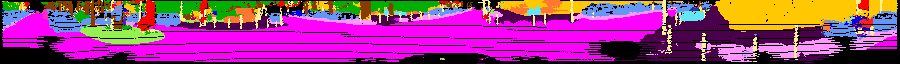}\\
Darknet53Seg\\ \includegraphics[width=\resultsize]{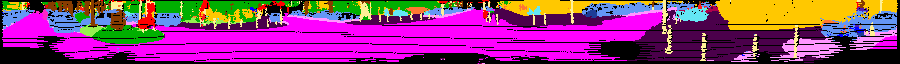}\\
Ground truth\\ \includegraphics[width=\resultsize]{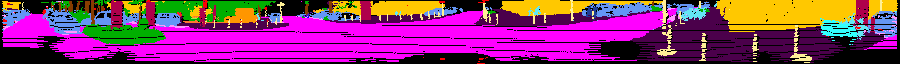}\\
\includegraphics[width=\resultsize]{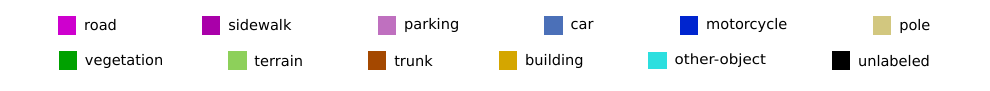} \\
\caption{Examples of inference for all methods. The point clouds
were projected to 2D using a spherical projection to make the comparison easier.}
% seq 13, scan 2877
\label{fig:results}
\end{figure*}

\input{script/table_result_multiscan_complete.tex}

\input{table_result_ssc.tex}

\section{Baseline Setup}

We modified the available implementations such that the methods could be trained and evaluated on our large-scale dataset with very sparse point clouds due to the LiDAR sensor. 
Note that most of these approaches have so far only been evaluated on small RGB-D indoor datasets.

We restricted the number of points within a single scan due to memory limitations on some approaches~\cite{qi2017cvpr,qi2017nips} to \num{50000} via random sampling. 

For SPLATNet, we used the $\text{SPLATNet}_{3D}$\footnote{\url{https://github.com/NVlabs/splatnet}} architecture from \cite{su2018cvpr}. The input consisted per point of the 3D position and its normal. The normals were previously estimated given \num{30} closest neighbors.

With TangentConv\footnote{\url{https://github.com/tatarchm/tangent_conv}} we used the existing configuration for Semantic3D. 
We sped up the training and validation procedures by precomputing scan batches and added asynchronous data loading. Complete single scans were provided during training. 
In the multi scan experiment we fixed the number of points per batch to \num{500\,000} due to memory constraints and started training from the single scan weights.

For SqueezeSeg~\cite{wu2018icra-scnn} and its Darknet backbone equivalents, we used a spherical projection of the
scans in the same way as the original SqueezeSeg approach. The projection contains \num{64} lines in height 
corresponding with the separate beams of the sensor, and extrapolating the configuration of SqueezeSeg which only
uses the front $90^\circ$ and a horizontal resolution of \num{512}, we use \num{2048} for the entire scan. Because
some points are duplicated in this sampling process, we always keep the closest range value, and in inference of
each scan we iterate over the entire point list and check it's semantic value in the output grid.

An overview of the used parameters is given in \reftab{tab:approach_stats}.
We furthermore provide the number of trained epochs and if we could get a results which seems to be converged in the given amount of time.

\section{Results using Multiple Scans}

The full per class IoU results for the multiple scans experiment are listed in \reftab{tab:results_multiscan_complete}.
As already mentioned in the main text, we generally observe that the IoU of static classes is mostly unaffected by the availability of multiple past scans.
To some extent, the IoU for some classes increases slightly.
The drop in performance in terms of mIoU is mainly caused by the additional challenge to correctly separate moving and non-moving classes.

\section{Semantic Scene Completion}

\reftab{tab:results_ssc} shows the class-wise results for semantic scene completion as well as precision and recall for scene completion. 
One can see that TS3D + DarkNet53Seg performs slightly better than SSCNet and TS3D. Note that DarkNet53Seg has been pretrained on the exact same classes as required for semantic scene completion. 
TS3D on the other hand uses DeepLab v2 (ResNet-101) \cite{chen2018pami} pretrained on the Cityscapes \cite{cordts2016cvpr} dataset, which does not differentiate between classes such as other-ground, parking or trunk for example. Another reason might be that 2D semantic labels projected back onto the point cloud is not very accurate especially at object boundaries, where labels often bleed onto distant objects. This is because in the 2D projection, they are close to each other, a problem that is inherent to the projection method. 
The best approach (TS3D + DarkNet53Seg + SATNet) outperforms the other approaches significantly (+20.77\% IoU on scene completion and +7.51\% mIoU on semantic scene completion). As mentioned above, it is the only approach capable of producing high resolution outputs. This approach however suffers from huge memory consumption. Therefore, during training the input volume is randomly cropped to volumes of grid size $64\times64\times32$ while during inference, each volume gets divided into 6 overlapping blocks of size $90\times138\times32$ for which the inference is performed individually. The individual blocks are subsequently fused to obtain the final result. Figure~\ref{fig:ssc_results} shows an example result of this approach.

Rare classes like bicycle, motorcycle, motorcyclist, and person are not or almost not recognized. This suggests that these classes are potentially hard to recognize, as they represent a small and rare signal in the \datasetname data.

\section{Qualitative Results}

\reffig{fig:results} shows qualitative results for the evaluated baseline approaches on a scan from the validation data.
Here we show the spherical projections of the results to enable an easier comparison of the results.

With increasing performance in terms of mean IoU (top to bottom), see also Table 2 of the paper, we see that ground points get better separated into the classes sidewalk, road, and parking.
In particular, parking areas need a lot of contextual information and also information from neighboring points, since often a small curb distinguishes the parking area from the road.

In general, one can see definitely an increased accuracy for smaller objects like the poles on the right side of the image, which indicates that the extra parameters of the models with the largest capacity ($25$ million as in the case of DarkNet21Seg and $50$ million as in the case of Darknet53Seg) are needed to distinguish smaller classes and class with few examples.

\section{Dataset and Baseline Access API}
Along with the annotations and the labeling tool, we also provide a public API implemented in Python. 

In contrast to our labeling tool, which is intended for allowing users to easily extend this dataset, and generate others for other purposes, this API  is intended to be used to easily access the data, calculate statistics, evaluate metrics, and access several implementations of different state-of-the-art semantic segmentation approaches. 
We hope that this API will serve as a baseline to implement new point cloud semantic segmentation approaches, and will provide a common framework to evaluate them, and compare them more transparently with other methods. 
The choice of Python as the underlying language for the API is that it is the current language of choice for the front end for deep learning framework developers, and therefore, for deep learning practitioners.

\begin{figure*}[t]
\centering
\includegraphics[width=0.95\linewidth]{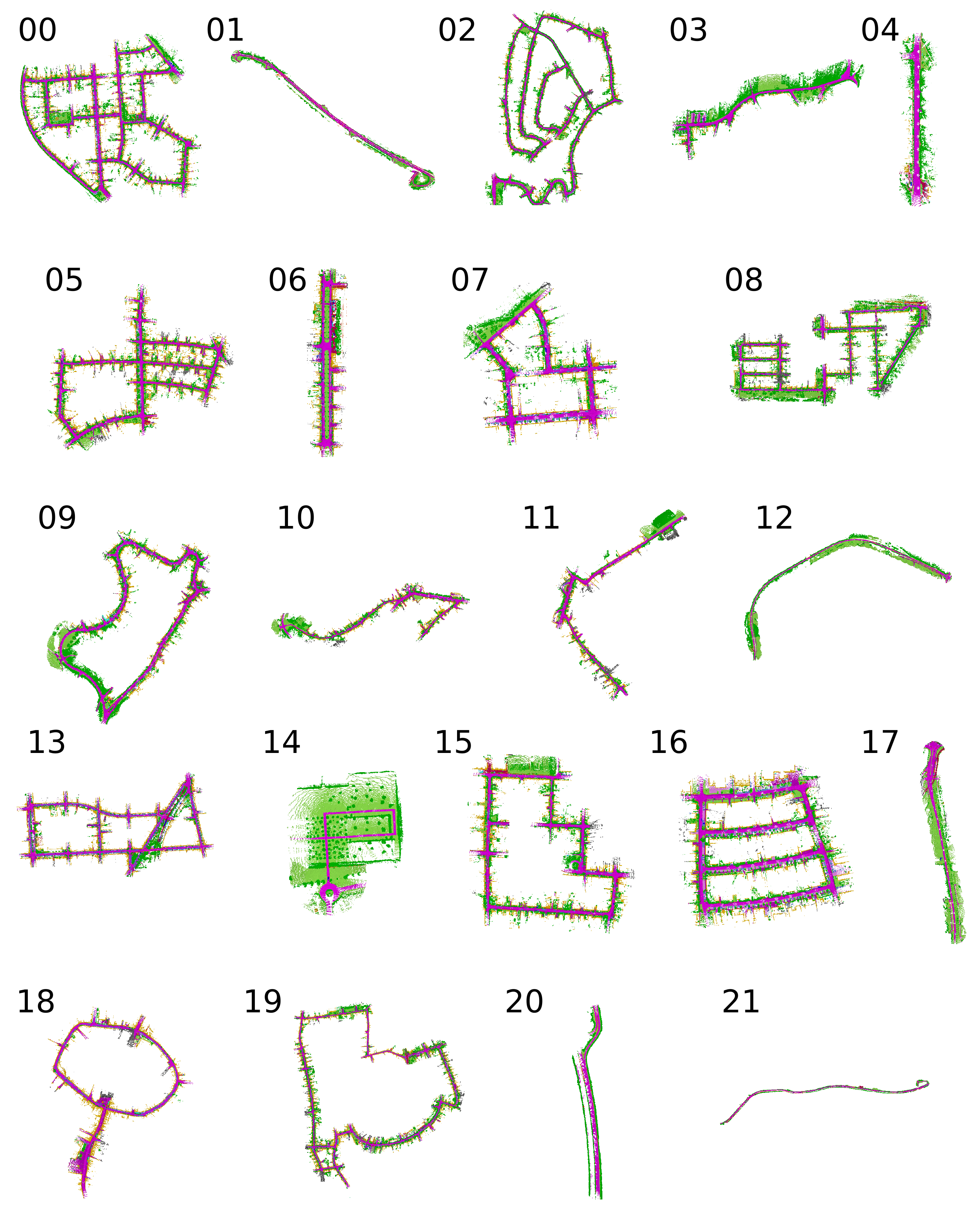}
\caption{Qualitative overview of labeled sequences and trajectories.}
\label{fig:overview}
\end{figure*}

\reffig{fig:overview} gives an overview of the labeled sequences showing the estimated trajectories and the aggregated point cloud over the whole sequence.

\end{document}

%% file: script/table_result_seg.tex
\setlength\tabcolsep{3.5pt}\begin{table*}[ht]\centering\footnotesize{
\begin{tabular}{lc|ccccccccccccccccccc}
\toprule 
Approach & \begin{sideways}mIoU\end{sideways} & \begin{sideways}road\end{sideways} & \begin{sideways}sidewalk\end{sideways} & \begin{sideways}parking\end{sideways} & \begin{sideways}other-ground\end{sideways} & \begin{sideways}building\end{sideways} & \begin{sideways}car\end{sideways} & \begin{sideways}truck\end{sideways} & \begin{sideways}bicycle\end{sideways} & \begin{sideways}motorcycle\end{sideways} & \begin{sideways}other-vehicle\end{sideways} & \begin{sideways}vegetation\end{sideways} & \begin{sideways}trunk\end{sideways} & \begin{sideways}terrain\end{sideways} & \begin{sideways}person\end{sideways} & \begin{sideways}bicyclist\end{sideways} & \begin{sideways}motorcyclist\end{sideways} & \begin{sideways}fence\end{sideways} & \begin{sideways}pole\end{sideways} & \begin{sideways}traffic sign\end{sideways}\\
\midrule
PointNet \cite{qi2017cvpr}  & 14.6 & 61.6 & 35.7 & 15.8 & 1.4 & 41.4 & 46.3 & 0.1 & 1.3 & 0.3 & 0.8 & 31.0 & 4.6 & 17.6 & 0.2 & 0.2 & 0.0 & 12.9 & 2.4 & 3.7\\
SPGraph \cite{landrieu2018cvpr}  & 17.4 & 45.0 & 28.5 & 0.6 & 0.6 & 64.3 & 49.3 & 0.1 & 0.2 & 0.2 & 0.8 & 48.9 & 27.2 & 24.6 & 0.3 & 2.7 & 0.1 & 20.8 & 15.9 & 0.8\\
SPLATNet \cite{su2018cvpr}  & 18.4 & 64.6 & 39.1 & 0.4 & 0.0 & 58.3 & 58.2 & 0.0 & 0.0 & 0.0 & 0.0 & 71.1 & 9.9 & 19.3 & 0.0 & 0.0 & 0.0 & 23.1 & 5.6 & 0.0\\
PointNet++ \cite{qi2017nips}  & 20.1 & 72.0 & 41.8 & 18.7 & 5.6 & 62.3 & 53.7 & 0.9 & 1.9 & 0.2 & 0.2 & 46.5 & 13.8 & 30.0 & 0.9 & 1.0 & 0.0 & 16.9 & 6.0 & 8.9\\
SqueezeSeg \cite{wu2018icra-scnn}  & 29.5 & 85.4 & 54.3 & 26.9 & 4.5 & 57.4 & 68.8 & 3.3 & 16.0 & 4.1 & 3.6 & 60.0 & 24.3 & 53.7 & 12.9 & 13.1 & 0.9 & 29.0 & 17.5 & 24.5\\
SqueezeSegV2 \cite{wu2019icra}  & 39.7 & 88.6 & 67.6 & 45.8 & 17.7 & 73.7 & 81.8 & 13.4 & 18.5 & 17.9 & 14.0 & 71.8 & 35.8 & 60.2 & 20.1 & 25.1 & 3.9 & 41.1 & 20.2 & 36.3\\
TangentConv \cite{tatarchenko2018cvpr}  & 40.9 & 83.9 & 63.9 & 33.4 & 15.4 & 83.4 & 90.8 & 15.2 & 2.7 & 16.5 & 12.1 & 79.5 & 49.3 & 58.1 & 23.0 & 28.4 & 8.1 & 49.0 & 35.8 & 28.5\\
DarkNet21Seg  & 47.4 & 91.4 & 74.0 & 57.0 & 26.4 & 81.9 & 85.4 & 18.6 & 26.2 & 26.5 & 15.6 & 77.6 & 48.4 & 63.6 & 31.8 & 33.6 & 4.0 & 52.3 & 36.0 & 50.0\\
DarkNet53Seg  & 49.9 & 91.8 & 74.6 & 64.8 & 27.9 & 84.1 & 86.4 & 25.5 & 24.5 & 32.7 & 22.6 & 78.3 & 50.1 & 64.0 & 36.2 & 33.6 & 4.7 & 55.0 & 38.9 & 52.2\\
\bottomrule
\end{tabular}}
\setlength\tabcolsep{6.0pt}\caption{Single scan results (19 classes) for all baselines on sequences 11 to 21 (test set). All methods were trained on sequences 00 to 10, except for sequence 08 which is used as validation set.}
\label{tab:results_segmentation}
\end{table*}

%% file: script/table_result_multiscan.tex
\bgroup
\setlength\tabcolsep{4.5pt}\begin{table}[t]\centering\footnotesize{
\begin{tabular}{lcccccc|c}
\toprule 
Approach & \begin{sideways}car\end{sideways} & \begin{sideways}truck\end{sideways} & \begin{sideways}other-vehicle\end{sideways} & \begin{sideways}person\end{sideways} & \begin{sideways}bicyclist\end{sideways} & \begin{sideways}motorcyclist\end{sideways} & \begin{sideways}mIoU\end{sideways}\\
\midrule
\multirow{2}{*}{TangentConv \cite{tatarchenko2018cvpr}}  & 84.9 & 21.1 & 18.5 & 1.6 & 0.0 & 0.0 & \multirow{2}{*}{34.1}\\
 & \cellcolor{Gainsboro!90} 40.3 & \cellcolor{Gainsboro!90} 42.2 & \cellcolor{Gainsboro!90} 30.1 & \cellcolor{Gainsboro!90} 6.4 & \cellcolor{Gainsboro!90} 1.1 & \cellcolor{Gainsboro!90} 1.9 & \\
 \rule{0pt}{0.5cm}\multirow{2}{*}{DarkNet53Seg}  & 84.1 & 20.0 & 20.7 & 7.5 & 0.0 & 0.0 & \multirow{2}{*}{41.6}\\
 & \cellcolor{Gainsboro!90} 61.5 & \cellcolor{Gainsboro!90} 37.8 & \cellcolor{Gainsboro!90} 28.9 & \cellcolor{Gainsboro!90} 15.2 & \cellcolor{Gainsboro!90} 14.1 & \cellcolor{Gainsboro!90} 0.2 & \\
\bottomrule
\end{tabular}}
\setlength\tabcolsep{6.0pt}\caption{IoU results using a sequence of multiple past scans (in \%). Shaded cells correspond to the IoU of the moving classes, while unshaded entries are the non-moving classes.}
\label{tab:results_multiscan}\vspace{-0.2cm}
\end{table}
\egroup

%% file: script/table_result_multiscan_complete.tex
\setlength\tabcolsep{1.9pt}\begin{table*}[ht]\centering\footnotesize{
\begin{tabular}{lccccccccccccccccccccccccc|c}
\toprule 
Approach & \begin{sideways}road\end{sideways} & \begin{sideways}sidewalk\end{sideways} & \begin{sideways}parking\end{sideways} & \begin{sideways}other-ground\end{sideways} & \begin{sideways}building\end{sideways} & \begin{sideways}car\end{sideways} & \begin{sideways}car (moving)\end{sideways} & \begin{sideways}truck\end{sideways} & \begin{sideways}truck (moving)\end{sideways} & \begin{sideways}bicycle\end{sideways} & \begin{sideways}motorcycle\end{sideways} & \begin{sideways}other-vehicle\end{sideways} & \begin{sideways}other-vehicle (moving)\end{sideways} & \begin{sideways}vegetation\end{sideways} & \begin{sideways}trunk\end{sideways} & \begin{sideways}terrain\end{sideways} & \begin{sideways}person\end{sideways} & \begin{sideways}person (moving)\end{sideways} & \begin{sideways}bicyclist\end{sideways} & \begin{sideways}bicyclist (moving)\end{sideways} & \begin{sideways}motorcyclist\end{sideways} & \begin{sideways}motorcyclist (moving)\end{sideways} & \begin{sideways}fence\end{sideways} & \begin{sideways}pole\end{sideways} & \begin{sideways}traffic-sign\end{sideways} & \begin{sideways}mIoU\end{sideways}\\
\midrule
TangentConv  & 83.9 & 64.0 & 38.3 & 15.3 & 85.8 & 84.9 & 40.3 & 21.1 & 42.2 & 2.0 & 18.2 & 18.5 & 30.1 & 79.5 & 43.2 & 56.7 & 1.6 & 6.4 & 0.0 & 1.1 & 0.0 & 1.9 & 49.1 & 36.4 & 31.2 & 34.1\\
DarkNet53Seg  & 91.6 & 75.3 & 64.9 & 27.5 & 85.2 & 84.1 & 61.5 & 20.0 & 37.8 & 30.4 & 32.9 & 20.7 & 28.9 & 78.4 & 50.7 & 64.8 & 7.5 & 15.2 & 0.0 & 14.1 & 0.0 & 0.2 & 56.5 & 38.1 & 53.3 & 41.6\\
\bottomrule
\end{tabular}}
\setlength\tabcolsep{6.0pt}\caption{IoU results using a sequence of multiple past scans (in \%).}
\label{tab:results_multiscan_complete}
\end{table*}

%% file: table_result_ssc.tex
\setlength\tabcolsep{3pt}\begin{table*}[ht]
\centering
\resizebox{\textwidth}{!}{
\footnotesize{
\begin{tabular}{lccc|ccccccccccccccccccc|c}
\toprule 

 & \multicolumn{3} {c|} {Scene Completion} & \multicolumn{19} {c|} {Semantic Scene Completion}  \\
Approach & \begin{sideways}precision\end{sideways} & \begin{sideways}recall\end{sideways} & \begin{sideways}IoU\end{sideways} & \begin{sideways}road\end{sideways} & \begin{sideways}sidewalk\end{sideways} & \begin{sideways}parking\end{sideways} & \begin{sideways}other-ground\end{sideways} & \begin{sideways}building\end{sideways} & \begin{sideways}car\end{sideways} & \begin{sideways}truck\end{sideways} & \begin{sideways}bicycle\end{sideways} & \begin{sideways}motorcycle\end{sideways} & \begin{sideways}other-vehicle\end{sideways} & \begin{sideways}vegetation\end{sideways} & \begin{sideways}trunk\end{sideways} & \begin{sideways}terrain\end{sideways} & \begin{sideways}person\end{sideways} & \begin{sideways}bicyclist\end{sideways} & \begin{sideways}motorcyclist\end{sideways} & \begin{sideways}fence\end{sideways} & \begin{sideways}pole\end{sideways} & \begin{sideways}traffic sign\end{sideways} & \begin{sideways}mIoU\end{sideways}\\
\midrule
SSCNet          & 31.71 & 83.40 & 29.83 & 27.55 & 16.99 & 15.60 & 6.04 & 20.88 & 10.35 & 1.79 & 0 & 0 & 0.11 & 25.77 & 11.88 & 18.16 & 0 & 0 & 0 & 14.40 & 7.90  & 3.67 & 9.53  \\
TS3D      & 31.58 & 84.18 & 29.81 & 28.00 & 16.98 & 15.65 & 4.86 & 23.19 & 10.72 & 2.39 & 0 & 0 & 0.19 & 24.73 & 12.46 & 18.32 & 0.03 & 0.05 & 0 & 13.23 & 6.98  & 3.52 & 9.54  \\
TS3D & & & & & & & & & & & & & & & & & & & & & &\\
+ DarkNet53Seg     & 25.85 & 88.25 & 24.99 & 27.53 & 18.51 & 18.89 & 6.58 & 22.05 & 8.04  & 2.19 & 0.08 & 0.02 & 3.96 & 19.48 & 12.85 & 20.22 & 2.33 & 0.61 & 0.01 & 15.79 & 7.57  & 6.99 & 10.19 \\
TS3D & & & & & & & & & & & & & & & & & & & & & &\\
+ DarkNet53Seg & & & & & & & & & & & & & & & & & & & & & &\\
+ SATNet & 80.52 & 57.65 & 50.60  & 62.20 & 31.57 & 23.29 & 6.46 & 34.12 & 30.70 & 4.85 & 0 & 0 & 0.07 & 40.12 & 21.88 & 33.09 & 0 & 0 & 0 & 24.05 & 16.89 & 6.94 & 17.70 \\
\bottomrule
\end{tabular}}
}
\setlength\tabcolsep{6.0pt}\caption{Results for scene completion and class-wise results for semantic scene completion (in \%).}
\label{tab:results_ssc}
\end{table*}